\pdfoutput=1

\documentclass[11pt]{article}

\usepackage[final]{acl}
\usepackage{listings}
\usepackage{times}
\usepackage{latexsym}
\usepackage{booktabs}
\usepackage[utf8]{inputenc}
\usepackage{xcolor}
\usepackage{boxedminipage}
\usepackage{subfig}

\usepackage{amsmath}
\usepackage{multirow}
\usepackage[T1]{fontenc}

\usepackage[utf8]{inputenc}

\usepackage{microtype}

\usepackage{inconsolata}

\usepackage{graphicx}
\usepackage{threeparttable} 
\usepackage[table]{xcolor}

\definecolor{fpr-high}{RGB}{255, 220, 220}   
\definecolor{fpr-mid} {RGB}{255, 235, 200}   
\definecolor{dir-down}{RGB}{220, 240, 220}   
\definecolor{dir-up}  {RGB}{255, 220, 220}   

\title{Style as a Confound: False Positives in AI Detection of Non-Native Academic Writing}

\author{
Hyeonchu Park\textsuperscript{1}, Gahye Jeong\textsuperscript{2}, Bugeun Kim\textsuperscript{1}\\
Department of Artificial Intelligence, Chung-Ang University, Republic of Korea\textsuperscript{1}\\
Wordvice, Republic of Korea\textsuperscript{2}\\
\{phchu0429, bgnkim\}@cau.ac.kr\textsuperscript{1}, gjeong@wordvice.com\textsuperscript{2}
}

\begin{document}
\maketitle
\begin{abstract}

AI text detectors are increasingly employed in academic settings, but it remains unclear whether their outputs reflect AI authorship itself or broader linguistic features associated with polished academic English. Previous studies have reported high false-positive rates (FPRs) for non-native English writing, but population-level comparisons confound authorship with differences in topic, domain, and writing style. Professional editing provides a useful setting for examining this issue because it changes the linguistic form of manuscripts while preserving authorship and content. We examined 135,389 document pairs from a professional English editing service (2018–2025), comprising non-native manuscripts and their native-edited versions, to assess how editing affects detector responses controlling for content and authorship. For the 13 AI text detectors, FPRs for human-written texts varied widely, from 0.0\% to 100.0\%. Responses varied across detectors: the same edits increased AI scores in some detectors but decreased them in others. Notably, score changes correlated with the extent of editing. 
The findings identify professional editing style as a key confounding variable in AI detector outputs, rather than establishing a full separation of text origin from linguistic style, raising concerns about fairness and reliability in academic settings.

\end{abstract}

\section{Introduction}


The rise of large language models(LLMs) has increased the use of AI text detectors in academia. While intended to uphold academic integrity, these systems often misclassify human-written text as AI-generated, raising concerns about false positives.

Recent studies have found that English text written by non-native speakers is more likely to be labeled as AI-generated text(AIGT) than text written by native speakers~\cite{non}. This finding suggests that linguistic properties (e.g., fluency and lexical diversity) may affect detector results, rather than the origin alone. However, prior work has focused on comparing writer populations, which are challenging to interpret because they are confounded by differences in topics, domains, writing purposes, and individual author characteristics.

This work addresses this limitation by treating professional academic English editing as a natural experiment. We compare original and professionally edited versions of manuscripts, holding content, topic, and authorship constant, and observe only the effects of linguistic revision. We evaluate 135,389 document pairs from a professional academic editing service and 13 representative AI text detectors. We examine (1) baseline FPRs on human-written academic texts, (2) effects of professional editing on detector results, (3) relationships between editing intensity and detector responses, and (4) whether these patterns differ between the pre-ChatGPT (2018–2022) and post-ChatGPT (2023–2025) periods.

The results demonstrate variation among detectors: some assign lower AI scores after professional editing, whereas others assign higher scores to the same text, and score shifts correlate with editing intensity. However, these shifts reach conventional thresholds for practical significance in only 7 of the 13 detectors, indicating that statistical significance and practical relevance diverge across the detector landscape. Using a large-scale paired-document design, this study provides evidence that detector outputs are sensitive to linguistic style even when authorship and content are fixed, raising fairness and reliability concerns for affected detectors.

\section{Related Work}
\label{sec:related}

\subsection{Non-Native Writing Bias}

Recent studies indicate that AI text detectors exhibit systematic variation in performance across writer populations, with non-native English speakers often facing higher false-positive rates than native speakers ~\cite{non}. This issue arises across various datasets, such as TOEFL essays and academic writing, where a significant amount of non-native speaker text is misclassified as AI-generated ~\cite{non, testing, problem}.

These disparities are linked to common features in non-native writing, including lower lexical diversity, simpler syntax, and increased repetition~\cite{corpus, pred, relation}. Many of these traits resemble those of AIGT~\cite{non, current}, raising concerns that detectors may focus on surface patterns rather than the text's true origins.

However, existing research mostly compares different writer groups, which limits understanding of the linguistic factors driving detector behavior and complicates analyses due to confounding variables like topic and writing style. As a result, higher detector scores for non-native writing may reflect differences in subject matter, disciplinary conventions, or individual writing habits rather than the effect of linguistic proficiency itself. In this work, we adopt a different perspective by examining how detector outputs vary when the same document is professionally edited. 

By comparing the original and edited versions, we aim to isolate the effects of linguistic refinement while keeping content and authorship constant, allowing for a clearer assessment of detector responses to text characteristics rather than origin.

\subsection{AI-Generated Text Detection}

Methods for detecting AIGT can be grouped into three broad categories: token-statistic, zero-shot, and classifier approaches.

\paragraph{Token-statistics-based detectors.}
Early approaches exploit statistical properties of token distributions produced by language models. Representative examples include Log-Likelihood, Log-Rank, Entropy, and GLTR~\cite{gltr}, which rely on measures such as token probability, token rank, and prediction entropy. These methods are motivated by the observation that machine-generated text often exhibits more predictable token distributions than human writing.

\paragraph{Zero-shot detectors.}
More recent methods infer text origin directly from language model behavior, without training dedicated classifiers. Some notable examples include DetectGPT~\cite{detectgpt}, Fast-DetectGPT~\cite{fast_detect}, DetectLLM-LRR ~\cite{detectllm}, LastDE+~\cite{lastde}, Binoculars~\cite{bino}, and DiVeye~\cite{div}. These methods use various signals, including probability curvature, likelihood ratios, entropy differences, and comparisons of probabilities across different models. Because they rely on language-model probabilities rather than task-specific training data, zero-shot detectors are often considered more adaptable to unseen generators and domains. However, their predictions still depend on linguistic properties reflected in model likelihoods, raising the possibility that writing proficiency and stylistic refinement may influence detector outputs.

\paragraph{Classifier-based detectors.}
A third family consists of supervised classifiers trained on large collections of human-written and AIGT. Representative examples include RoBERTa-based detectors~\cite{roberta}, MAGE~\cite{mage}, and RADAR~\cite{radar}. These models learn decision boundaries from labeled training data and often achieve strong benchmark performance. Because their predictions are derived from patterns learned during training, classifier-based detectors may capture a broad mixture of signals related to both text origin and writing style. As a result, linguistic characteristics commonly associated with non-native writing may inadvertently become predictive cues during training, potentially contributing to the disparities reported in prior work.

Although these detector families rely on different underlying principles, most evaluations have been conducted in benchmark settings that compare human-written and AI-generated corpora. Consequently, it remains unclear whether detector outputs primarily reflect the origin of generation or broader linguistic properties that happen to correlate with AIGT. This uncertainty is particularly important in light of the non-native writing bias discussed above, since linguistic features associated with writing proficiency may be mistaken for evidence of AI generation. Understanding which detector paradigms are most sensitive to such stylistic factors is therefore critical for evaluating the fairness and reliability of AI text detection systems.

This work addresses this question by comparing detector results across three detection paradigms using professionally edited document pairs. This setting permits the assessment of how detector results change in response to controlled linguistic refinement while holding document content and authorship constant.

\section{Method}
\label{sec:method}

\subsection{Data Source}




We employed editing-record data from a professional academic English editing service\footnote{\url{https://wordvice.ai/}}. Each record comprises two versions of a document: the original text written by a non-native English speaker and the version professionally edited by a native-speaker editor.
 
\paragraph{Editorial protocol.}
All editors follow a standardized internal editing-guidelines document (25 pages) that governs the editing process. The protocol is organized around three principles: (1) \textit{meaning preservation} — editors correct grammar, punctuation, mechanics, clarity, naturalness, and word choice while preserving the author's meaning; (2) \textit{human editorial judgment} — stylistic revision, vocabulary selection, and sentence restructuring are performed exclusively by the human editor based on expertise, not by automated rewriting; and (3) \textit{restricted AI usage} — company policy permits AI-based grammar-checking tools only as a supplementary aid for catching objective mechanical errors (e.g., subject-verb agreement, punctuation), while AI-generated rewriting, paraphrasing, or stylistic generation is explicitly prohibited.
 
\paragraph{Dataset statistics.}
The final dataset consists of 135,389 original--edited document pairs collected between 2018 and 2025, spanning more than 40 academic disciplines. Table~\ref{tab:dataset-summary} summarizes the key statistics; the complete breakdown by year, service type, subject domain, and English variant is provided in Appendix~\ref{app:dataset}.
 
\begin{table}[t]
\centering
\small
\begin{tabular}{lr}
\toprule
Statistic & Value \\
\midrule
Total document pairs        & 135{,}389 \\
Mean word count (original)  & 2{,}355 \\  
Median word count (original)& 950 \\      
Std.\ dev.\ (word count)    & 3{,}038 \\
Subject areas                & 40+ \\
Docs.\ w/ 500--5{,}000 words & $\sim$57\% \\
\bottomrule
\end{tabular}
\caption{Summary dataset statistics. Full year-by-year and service-type breakdown in Appendix~\ref{app:dataset}}
\label{tab:dataset-summary}
\end{table}
 
\paragraph{Characteristics of professional edits.}
To characterize the edits themselves (independent of any detector), we compared surface linguistic properties of original and edited texts on a randomly sampled subset of 6,000 document pairs (Table~\ref{tab:edit-characteristics}).
Professional editing introduces modest lexical and readability changes while largely preserving document length and structure: median document length changed by only 6.5 words. Editing intensity, measured in edit units, has a median of 196 (mean 463) per document, indicating that the edits are substantial enough to meaningfully affect wording while leaving overall content intact.
 
\begin{table}[t]
\centering
\small
\begin{tabular}{lrrr}
\toprule
Metric & Orig. & Edited & $\Delta$ \\
\midrule
Mean word length      & 5.196 & 5.252 & $+0.055$ \\
Type--token ratio      & 0.407 & 0.412 & $+0.005$ \\
Mean sentence length   & 24.35 & 23.51 & $-0.84$ \\
Flesch--Kincaid grade  & 14.51 & 14.42 & $-0.09$ \\
\bottomrule
\end{tabular}
\caption{Linguistic characteristics of professional editing
(6{,}000-pair subsample). Edit intensity: median 196 / mean 463 edit
units per document.}
\label{tab:edit-characteristics}
\end{table}

\subsection{AI Detectors}

We evaluate 13 publicly available AI text detectors representing diverse detection paradigms. These detectors are categorized into three groups:

\paragraph{Token-statistics-based}
Log-rank~\cite{gltr}, Log-likelihood~\cite{likeli}, Entropy~\cite{gltr}, and GLTR~\cite{gltr}: These methods rely on statistical properties of token distributions, such as predictability, token rank, and entropy.

\paragraph{Zero-shot-based}
Fast-DetectGPT~\cite{fast_detect}, LastDE+~\cite{lastde}, DetectLLM-LRR~\cite{detectllm}, DiVeye~\cite{div}, BiScope~\cite{biscope}, Binoculars~\cite{bino}: These approaches estimate the likelihood of AI generation using language model probabilities or likelihood ratios across different language models.

\paragraph{Classifier-based}
RoBERTa~\cite{roberta}, MAGE~\cite{mage}, RADAR~\cite{radar}: These detectors learn decision boundaries through supervised training on human-written and AIGT corpora.

We evaluated all detectors in their publicly available, off-the-shelf configurations, without detector-specific fine-tuning or additional training on our dataset. This wide coverage enables a comparative analysis of how different detection paradigms respond to non-native English writing and whether certain approaches are more vulnerable to false positives. Appendix~\ref{app:model} describes all assessed detectors.

\subsection{Experimental Design}
\subsubsection{Baseline False Positive Analysis}

We first evaluate the baseline false-positive behavior of each detector on the Pre-ChatGPT dataset. Since these documents were written before the widespread adoption of modern generative AI systems, they can be reasonably treated as human-written texts with negligible AI involvement. Consequently, any AI-generated classification is considered a false positive

For each detector, we computed the FPR, accuracy, and F1-score to characterize baseline detector performance. For paired analyses of original and edited documents, we additionally report the change in false-positive rate ($\Delta$FP), defined as the difference in FPR between the edited and original versions, to quantify how professional editing alters binary detector decisions.


\subsubsection{Effectiveness of Professional Editing}

Next, we investigate how professional editing affects detector outputs. Because each original-edited pair corresponds to the same document, differences in detection results can be attributed primarily to linguistic revisions rather than differences in topic, content, or authorship.

\begin{equation}
    \Delta\text{score} = score_{edit} - score_{orig}
\end{equation}

For analyses involving temporal trends, score shifts were computed separately for the pre-ChatGPT (2018--2022) and post-ChatGPT (2023--2025) subsets. This approach examines whether the effects of professional editing changed the detector results after the widespread adoption of LLM-assisted writing.

Negative values indicate that the detector assigns a lower AI score after editing, whereas positive values indicate an increase in AI likelihood.

If detectors primarily capture text origin, score shifts should remain relatively small because both versions are human-written. In contrast, if detectors are sensitive to linguistic fluency or stylistic sophistication, systematic changes in detector outputs may emerge following professional editing.

\subsubsection{Editing Intensity Analysis}

To investigate the potential mechanisms underlying detector responses, we measure editing intensity using the edit ratio:

\begin{equation}
r_{edit}
=
\frac{|C_{del}|+|C_{add}|}
{|C_{orig}|}
\end{equation}

where $C_{del}$, $C_{add}$, and $C_{orig}$ are the number of deleted, added, and initial characters, respectively, in the document.
To leverage information about the extent of human revision, we use the edit ratio rather than the normalized edit distance. Because normalized edit distance is based on the minimum number of editing operations required to transform one text into another, it may not accurately reflect the actual editing actions performed by professional editors. As a result, it can distort the magnitude of true editorial intervention. The edit ratio instead measures the volume of added and removed content, making it more suitable for quantifying editing intensity.

\subsection{Statistical Analysis}

Differences between original and edited texts are evaluated using paired-sample $t$-tests, with effect sizes reported as Cohen's $d$ and 95\% confidence intervals estimated via 1,000 bootstrap resamples. The paired design accounts for document-level variation by comparing each edited document directly with its original version.

We assess the relationship between editing intensity and shifts in detector scores using Spearman's rank correlations and linear regression models. Documents are additionally grouped into editing-intensity quintiles to examine dose--response patterns. Consistent monotonic trends across quintiles indicate systematic detector sensitivity to increasing levels of editorial intervention.

For temporal analyses, outcomes are evaluated separately for the pre-ChatGPT (2018-2022) and post-ChatGPT (2023-2025) periods. We compare detector score shifts and false-positive rates across periods to examine whether detector behavior changed following the widespread adoption of LLM-assisted writing. Bonferroni and Benjamini-Hochberg corrections are applied where appropriate to account for multiple hypothesis testing.

\section{Results}
\label{sec:results}
\subsection{False-Positive Results Vary by Detector}
\begin{table}[t]
\centering\small
\begin{tabular}{l rr rr}
\toprule
\multirow{2}{*}{\textbf{Detector}}
  & \multicolumn{2}{c}{\textbf{FP\%}}
  & \multirow{2}{*}{\textbf{$\Delta$FP}}
  & \multirow{2}{*}{\textbf{$p$}} \\
\cmidrule(lr){2-3}
  & orig & edit & & \\
\midrule
\multicolumn{5}{l}{\textit{Token-statistics-based}} \\
Log-rank
  & \cellcolor{fpr-high}100.0 & \cellcolor{fpr-high}100.0
  & $+$0.0 & ns \\
Log-likelihood
  & \cellcolor{fpr-high}99.8  & \cellcolor{fpr-high}99.9
  & $+$0.0 & $<$.05 \\
Entropy
  & \cellcolor{fpr-high}99.8  & \cellcolor{fpr-high}99.9
  & $+$0.1 & $<$.001 \\
GLTR
  & \cellcolor{fpr-high}99.9  & \cellcolor{fpr-high}99.9
  & $+$0.0 & $<$.01 \\
\midrule
\multicolumn{5}{l}{\textit{Zero-shot-based}} \\
Fast-DetectGPT
  & \cellcolor{fpr-mid}25.2 & \cellcolor{fpr-mid}31.6
  & \cellcolor{dir-up}$+$6.4 & $<$.001 \\
LastDE+
  & \cellcolor{fpr-mid}19.9 & \cellcolor{fpr-mid}27.3
  & \cellcolor{dir-up}$+$7.4 & $<$.001 \\
DetectLLM-LRR
  & 0.2 & 0.3
  & $+$0.1 & $<$.01 \\
DiVeye
  & 0.4 & 0.6
  & $+$0.2 & $<$.001 \\
BiScope
  & 0.6 & 0.3
  & \cellcolor{dir-down}$-$0.4 & $<$.001 \\
Binoculars
  & 0.0 & 0.1
  & $+$0.0 & ns \\
\midrule
\multicolumn{5}{l}{\textit{Classifier-based}} \\
RoBERTa
  & \cellcolor{fpr-high}93.1 & \cellcolor{fpr-high}93.9
  & \cellcolor{dir-up}$+$0.8 & $<$.001 \\
MAGE
  & \cellcolor{fpr-mid}16.7  & 7.6
  & \cellcolor{dir-down}$-$9.1 & $<$.001 \\
RADAR
  & \cellcolor{fpr-high}88.3 & \cellcolor{fpr-high}86.4
  & \cellcolor{dir-down}$-$1.9 & $<$.001 \\
\bottomrule
\end{tabular}
\caption{%
  Baseline false positive behavior on the Pre-ChatGPT dataset (2018--2022).
  FPR color: \colorbox{fpr-high}{$\geq$50\%}, \colorbox{fpr-mid}{10--49\%}.
  $\Delta$FP color: \colorbox{dir-down}{bias reduction}, \colorbox{dir-up}{bias amplification}.
  Full results in Appendix~\ref{app:rq2a_score}}
\label{tab:rq1_baseline}
\end{table}

We assess the baseline false-positive rates of AI detectors using a dataset from before the introduction of ChatGPT (2018–2022), assuming minimal influence from modern generative AI. Since all documents were created before the advent of LLMs, any classification indicating that a document is AI-generated is considered a false positive.

Table~\ref{tab:rq1_baseline} illustrates variability among the detectors, showing FPR that range from less than 1\% for some detectors to nearly 100\% for others when analyzing the same set of human-written academic texts. Detectors based on token statistics exhibited the highest false positive rates, often misclassifying the majority of documents as AI-generated. In contrast, zero-shot likelihood-based detectors maintained an FPR close to zero. Classifier-based methods presented mixed results. These inconsistencies suggest that detector performance is influenced by the methodology used: token predictability often mislabels non-native academic writing as AI-generated, while more conservative likelihood-ratio-based approaches tend to be more accurate.

Overall, the discrepancies among detectors raise questions about what they are actually measuring. The same human-written documents can be classified by different systems as either predominantly AI-generated or human-written. Detectors may rely on various cues that prompt further analysis of how outputs change with controlled linguistic revisions through professional editing.

\subsection{Editing Alters Detector Outputs}
\label{result2}
\begin{table*}[t]
\centering\small
\begin{tabular}{l l rr rr rr}
\toprule
& & \multicolumn{2}{c}{\textbf{ALL}}
  & \multicolumn{2}{c}{\textbf{Pre-ChatGPT}}
  & \multicolumn{2}{c}{\textbf{Post-ChatGPT}} \\
\cmidrule(lr){3-4}\cmidrule(lr){5-6}\cmidrule(lr){7-8}
\textbf{Detector} & \textbf{Type}
  & $\Delta\text{score}$ & $\Delta\text{FP}$
  & $\Delta\text{score}$ & $\Delta\text{FP}$
  & $\Delta\text{score}$ & $\Delta\text{FP}$ \\
\midrule

Entropy        & Stat.
  & \cellcolor{dir-up}$+$0.014$^{***}$ & \cellcolor{dir-up}$+$0.1
  & \cellcolor{dir-up}$+$0.017$^{***}$ & \cellcolor{dir-up}$+$0.1
  & \cellcolor{dir-up}$+$0.005$^{***}$ & $+$0.0 \\
GLTR           & Stat.
  & \cellcolor{dir-up}$+$0.010$^{***}$ & $+$0.0
  & \cellcolor{dir-up}$+$0.012$^{***}$ & $+$0.0
  & \cellcolor{dir-up}$+$0.004$^{***}$ & $+$0.0 \\
Log-rank       & Stat.
  & \cellcolor{dir-up}$+$0.001$^{ns}$  & $+$0.0
  & \cellcolor{dir-up}$+$0.001$^{***}$ & $+$0.0
  & \cellcolor{dir-up}$+$0.001$^{***}$ & $+$0.0 \\
Log-like       & Stat.
  & \cellcolor{dir-up}$+$0.005$^{ns}$  & $+$0.0
  & \cellcolor{dir-up}$+$0.006$^{***}$ & $+$0.0
  & \cellcolor{dir-up}$+$0.002$^{***}$ & $+$0.0 \\
\midrule

BiScope        & Zero
  & \cellcolor{dir-down}$-$0.019$^{***}$ & \cellcolor{dir-down}$-$0.5
  & \cellcolor{dir-down}$-$0.021$^{***}$ & \cellcolor{dir-down}$-$0.6
  & \cellcolor{dir-down}$-$0.008$^{***}$ & \cellcolor{dir-down}$-$0.1 \\
Binoculars     & Zero
  & $+$0.002$^{ns}$  & $+$0.0
  & $+$0.003$^{***}$ & $+$0.0
  & $+$0.000$^{ns}$  & $+$0.0 \\
DetectLLM-LRR  & Zero
  & \cellcolor{dir-up}$+$0.003$^{***}$ & \cellcolor{dir-up}$+$0.0
  & \cellcolor{dir-up}$+$0.003$^{***}$ & \cellcolor{dir-up}$+$0.0
  & \cellcolor{dir-up}$+$0.002$^{***}$ & \cellcolor{dir-up}$+$0.0 \\
DiVeye         & Zero
  & \cellcolor{dir-up}$+$0.011$^{***}$ & \cellcolor{dir-down}$-$1.0
  & \cellcolor{dir-up}$+$0.013$^{***}$ & \cellcolor{dir-down}$-$1.2
  & \cellcolor{dir-up}$+$0.005$^{ns}$  & \cellcolor{dir-down}$-$0.3 \\
Fast-DetectGPT & Zero
  & \cellcolor{dir-up}$+$0.032$^{***}$ & \cellcolor{dir-up}$+$4.6
  & \cellcolor{dir-up}$+$0.038$^{***}$ & \cellcolor{dir-up}$+$5.4
  & \cellcolor{dir-up}$+$0.013$^{***}$ & \cellcolor{dir-up}$+$1.6 \\
LastDE+        & Zero
  & \cellcolor{dir-up}$+$0.027$^{***}$ & \cellcolor{dir-up}$+$5.8
  & \cellcolor{dir-up}$+$0.031$^{***}$ & \cellcolor{dir-up}$+$6.9
  & \cellcolor{dir-up}$+$0.012$^{***}$ & \cellcolor{dir-up}$+$2.0 \\
\midrule

MAGE    & Cls.
  & \cellcolor{dir-down}$-$0.108$^{***}$ & \cellcolor{dir-down}$-$10.7
  & \cellcolor{dir-down}$-$0.130$^{***}$ & \cellcolor{dir-down}$-$13.0
  & \cellcolor{dir-down}$-$0.031$^{***}$ & \cellcolor{dir-down}$-$3.1 \\
RADAR   & Cls.
  & \cellcolor{dir-down}$-$0.037$^{***}$ & \cellcolor{dir-down}$-$4.7
  & \cellcolor{dir-down}$-$0.044$^{***}$ & \cellcolor{dir-down}$-$5.6
  & \cellcolor{dir-down}$-$0.013$^{***}$ & \cellcolor{dir-down}$-$1.5 \\
RoBERTa & Cls.
  & \cellcolor{dir-up}$+$0.017$^{***}$ & \cellcolor{dir-up}$+$1.7
  & \cellcolor{dir-up}$+$0.018$^{***}$ & \cellcolor{dir-up}$+$1.8
  & \cellcolor{dir-up}$+$0.014$^{***}$ & \cellcolor{dir-up}$+$1.3 \\
\bottomrule
\end{tabular}
\caption{%
  Effect of professional editing on AI detector scores and false positive rates.
  $\Delta\text{score} = \text{score}_{\text{edit}} - \text{score}_{\text{orig}}$ (mean score shift);
  $\Delta\text{FP} = \text{FP}_{\text{edit}} - \text{FP}_{\text{orig}}$.
  \colorbox{dir-down}{Green}\,=\,decrease; \colorbox{dir-up}{Red}\,=\,increase.
  $^{***}p<.001$, $^{**}p<.01$, $^{*}p<.05$, $^{ns}p\geq.05$.
  Full results in Appendix~\ref{app:rq2a_score}}
\label{tab:rq2_combined}
\end{table*}

We next examine how professional editing affects detector outputs. Because each original--edited pair corresponds to the same document, observed score differences can be attributed primarily to linguistic revisions rather than differences in topic, content, or authorship. The results are presented in Table ~\ref{tab:rq2_combined}, with full results available in Appendix~\ref{app:rq2a_score}.

Professional editing led to significant score shifts across nearly all detectors, with varying directions and magnitudes. Some detectors, such as MAGE ($\Delta$score = $-0.108$, $\Delta$FP = $-10.7$ pp), RADAR ($-0.037$, $-4.7$), and BiScope ($-0.019$, $-0.5$), were less likely to classify texts as AI-generated post-editing. Conversely, detectors such as Fast-DetectGPT and LastDE+ showed notable increases in false positive rates, with $\Delta$FP of $+4.6$ and $+5.8$ pp, respectively.

A notable pattern is that the same professional editing produced contradictory responses across detector categories. Classifier-based detectors (e.g., MAGE and RADAR) tended toward the human-written results, whereas most token-statistics-based detectors tended toward the AI-generated results. Zero-shot detectors displayed mixed results, with BiScope producing negative shifts and Fast-DetectGPT and LastDE+ generating positive shifts. Notably, these patterns were largely consistent across pre- and post-ChatGPT subsets. Although the effect magnitude was typically smaller in the post-ChatGPT period, detectors that became more permissive after editing in the full dataset tended to remain permissive, whereas detectors that became more restrictive continued to move in that direction.

The findings indicate that detector disagreement extends beyond baseline FPRs. Even with the same human revision, detectors yield different interpretations, suggesting reliance on distinct linguistic signals rather than a unified definition of AIGT.

\subsection{Responses Scale with Editing Intensity}
\label{result3}
\begin{table*}[t]
\centering\small
\begin{tabular}{l l r r r r r r r}
\toprule
\textbf{Detector} & \textbf{Type} & $\rho$ & $R^{2}$
  & Q1 & Q2 & Q3 & Q4 & Q5 \\
\midrule

Entropy        & Stat.
  & \cellcolor{dir-up}$+$0.343$^{***}$ & 0.037
  & \cellcolor{dir-down}$-$.009 & \cellcolor{dir-up}$+$.015 & \cellcolor{dir-up}$+$.020 & \cellcolor{dir-up}$+$.027 & \cellcolor{dir-up}$+$.032 \\
Log-rank       & Stat.
  & \cellcolor{dir-up}$+$0.293$^{***}$ & 0.012
  & \cellcolor{dir-up}$+$.001 & \cellcolor{dir-up}$+$.001 & \cellcolor{dir-up}$+$.002 & \cellcolor{dir-up}$+$.002 & \cellcolor{dir-up}$+$.002 \\
Log-likelihood & Stat.
  & \cellcolor{dir-up}$+$0.287$^{***}$ & 0.005
  & \cellcolor{dir-up}$+$.003 & \cellcolor{dir-up}$+$.005 & \cellcolor{dir-up}$+$.006 & \cellcolor{dir-up}$+$.007 & \cellcolor{dir-up}$+$.008 \\
GLTR           & Stat.
  & \cellcolor{dir-up}$+$0.257$^{***}$ & 0.023
  & \cellcolor{dir-up}$+$.006 & \cellcolor{dir-up}$+$.011 & \cellcolor{dir-up}$+$.013 & \cellcolor{dir-up}$+$.016 & \cellcolor{dir-up}$+$.019 \\
\midrule

BiScope        & Zero
  & \cellcolor{dir-down}$-$0.418$^{***}$ & 0.041
  & \cellcolor{dir-up}$+$.009 & \cellcolor{dir-down}$-$.019 & \cellcolor{dir-down}$-$.026 & \cellcolor{dir-down}$-$.033 & \cellcolor{dir-down}$-$.040 \\
Binoculars     & Zero
  & \cellcolor{dir-down}$-$0.134$^{***}$ & 0.019
  & \cellcolor{dir-up}$+$.028 & \cellcolor{dir-down}$-$.002 & \cellcolor{dir-down}$-$.003 & \cellcolor{dir-down}$-$.004 & \cellcolor{dir-down}$-$.004 \\
DetectLLM-LRR  & Zero
  & \cellcolor{dir-up}$+$0.178$^{***}$ & 0.013
  & \cellcolor{dir-up}$+$.002 & \cellcolor{dir-up}$+$.003 & \cellcolor{dir-up}$+$.003 & \cellcolor{dir-up}$+$.004 & \cellcolor{dir-up}$+$.005 \\
DiVeye         & Zero
  & \cellcolor{dir-up}$+$0.147$^{***}$ & 0.012
  & \cellcolor{dir-up}$+$.008 & \cellcolor{dir-up}$+$.012 & \cellcolor{dir-up}$+$.013 & \cellcolor{dir-up}$+$.016 & \cellcolor{dir-up}$+$.020 \\
Fast-DetectGPT & Zero
  & $+$0.026$^{***}$ & 0.001
  & \cellcolor{dir-up}$+$.025 & \cellcolor{dir-up}$+$.036 & \cellcolor{dir-up}$+$.032 & \cellcolor{dir-up}$+$.041 & \cellcolor{dir-up}$+$.055 \\
LastDE+        & Zero
  & $+$0.026$^{***}$ & 0.001
  & \cellcolor{dir-up}$+$.027 & \cellcolor{dir-up}$+$.030 & \cellcolor{dir-up}$+$.030 & \cellcolor{dir-up}$+$.032 & \cellcolor{dir-up}$+$.037 \\
\midrule

MAGE    & Cls.
  & \cellcolor{dir-down}$-$0.193$^{***}$ & 0.028
  & \cellcolor{dir-down}$-$.032 & \cellcolor{dir-down}$-$.093 & \cellcolor{dir-down}$-$.133 & \cellcolor{dir-down}$-$.184 & \cellcolor{dir-down}$-$.214 \\
RADAR   & Cls.
  & \cellcolor{dir-down}$-$0.143$^{***}$ & 0.033
  & \cellcolor{dir-down}$-$.014 & \cellcolor{dir-down}$-$.031 & \cellcolor{dir-down}$-$.038 & \cellcolor{dir-down}$-$.048 & \cellcolor{dir-down}$-$.058 \\
RoBERTa & Cls.
  & $+$0.029$^{***}$ & 0.003
  & \cellcolor{dir-up}$+$.012 & \cellcolor{dir-up}$+$.018 & \cellcolor{dir-up}$+$.020 & \cellcolor{dir-up}$+$.024 & \cellcolor{dir-up}$+$.028 \\
\bottomrule
\end{tabular}
\caption{%
  Editing intensity and detector score shifts.
  Spearman $\rho$ denotes the correlation between edit-token ratio and
  $\Delta\text{score}$.
  Q1--Q5 represent quintiles of editing intensity (Q1\,=\,weakest, Q5\,=\,strongest).
  \colorbox{dir-down}{Green}\,=\,score decreases; \colorbox{dir-up}{Red}\,=\,score increases.
  $^{***}p<.001$, $^{**}p<.01$, $^{*}p<.05$.}
\label{tab:rq3_dose}
\end{table*}

To understand the observed editing effects, we analyzed how detector responses correlate with the extent of editorial revision. If professional editing impacts detector outputs through linguistic refinement, detectors sensitive to editing should show greater score shifts with more extensive revisions.

Table~\ref{tab:rq3_dose} reveals that detectors with decreased AI scores after editing exhibited negative correlations between edit ratio and score shift, while those with increased scores showed positive correlations. Notable examples include MAGE and BiScope, which had the largest score reductions and strong negative associations with editing intensity ($\rho=-0.193$ and $\rho=-0.418$).

For MAGE, mean score shifts ranged from $-0.032$ in the lowest editing quintile (Q1) to $-0.214$ in the highest (Q5), indicating that more extensive revisions pushed documents closer to the human-written region. In contrast, detectors such as Entropy ($\rho=+0.343$) and Log-rank ($\rho=+0.293$) showed positive score shifts as editing intensity increased, suggesting that editing shifted texts toward the AI-generated region.

Editing intensity did not change the direction of detector responses but amplified existing patterns. Detectors that viewed professional editing as evidence of human authorship became more permissive with increased revisions, while those that interpreted editing as a sign of AI generation became more restrictive. Although the explanatory power of editing intensity was modest ($R^2 \leq 0.041$), consistent trends across detector families indicate that outputs are systematically influenced by linguistic modifications. These findings suggest that AI detectors respond not only to the origin of generation but also to the extent of linguistic refinement through professional editing. The same qualitative trends are observed when the analysis is restricted to documents with substantial editorial revisions (Appendix~\ref{app:ab2}).

\subsection{Linguistic Drivers of Detector Score Shifts}
\label{sec:linguistic-drivers}

The preceding results show that professional editing shifts detector
scores in opposite directions across detector families
(Section~\ref{result2}) and that these shifts scale with editing
intensity (Section~\ref{result3}). We next examine which linguistic
properties of the edits are associated with these shifts.

\paragraph{Correlations with detector score shifts.}
We correlate editing characteristics with $\Delta$score for
statistical, LLM-based zero-shot, and supervised detectors on the
pre-ChatGPT subset (Table~\ref{tab:family-correlations}).
Edit-token ratio shows the strongest association, with a positive
correlation for statistical detectors ($\rho=+0.286$) but a negative
correlation for supervised detectors ($\rho=-0.089$). Grammar and
syntactic ratios exhibit the same qualitative pattern.

\begin{table}[t]
\centering
\small
\begin{tabular}{lrrr}
\toprule
Feature & Statistical & LLM-based & Supervised \\
\midrule
Grammar ratio & $+0.111$ & $+0.019$ & $-0.005$ \\
Lexical ratio & $+0.042$ & $-0.018$ & $-0.020$ \\
Syntactic ratio & $+0.101$ & $+0.013$ & $-0.040$ \\
Edit-token ratio & $+0.286$ & $+0.022$ & $-0.089$ \\
Orig.\ word count & $-0.099$ & $-0.002$ & $+0.067$ \\
\# edit units & $+0.116$ & $+0.021$ & $+0.005$ \\
\bottomrule
\end{tabular}
\caption{Spearman correlation ($\rho$) between editing characteristics
and detector score shifts, by detector family (pre-ChatGPT subset).}
\label{tab:family-correlations}
\end{table}

We additionally examined changes in linguistic complexity on a
6,000-document subsample (Table~\ref{tab:complexity-correlations}).
Changes in function-word ratio show positive correlations across all
detector families, whereas changes in type--token ratio show negative
correlations. Associations with average word and sentence length are
weaker and less consistent. These results suggest that detector score
shifts are more closely associated with fluency- and syntax-related
changes than with semantic content.

\begin{table}[t]
\centering
\small
\setlength{\tabcolsep}{4pt}
\begin{tabular}{lrrr}
\toprule
Metric & Statistical & LLM-based & Supervised \\
\midrule
TTR                 & $-0.063$ & $-0.063$ & $-0.073$ \\
Word length         & $-0.022$ & $-0.046$ & $-0.082$ \\
Function-word ratio & $+0.071$ & $+0.068$ & $+0.064$ \\
Sentence length     & $-0.043$ & $+0.005$ & $+0.017$ \\
\bottomrule
\end{tabular}
\caption{Spearman correlation between changes in complexity and
detector score shifts, by detector family.}
\label{tab:complexity-correlations}
\end{table}

\paragraph{Interpretation.}
Together with the practical-significance analysis in Section~\ref{result2},
these results suggest systematic differences across detector families.
Statistical detectors show increasing AI-likelihood with greater
editing volume, potentially because professional editing reduces
token-level perplexity and increases linguistic predictability.
Supervised classifiers show the opposite trend, consistent with their
reliance on higher-level patterns learned from AI-generated text.
LLM-based zero-shot detectors exhibit comparatively weak and mixed
relationships. These interpretations are correlational and therefore
do not establish causal effects of any individual linguistic feature.

\section{Conclusion}

This study examined the impact of professional academic editing on AI text-detection behavior using 135,389 pairs of original and edited documents from an academic editing platform. By comparing pre- and post-editing versions, we isolated the effects of linguistic changes while controlling for topic, content, and authorship.

Results showed significant variability among detectors, with false-positive rates differing widely for the same human-written texts. Editing affected AI-generation scores in varying ways, with some detectors showing lower scores post-edit and others showing higher scores for the same edits. There was also a correlation between detector responses and editing intensity, suggesting a link to linguistic revision rather than random variation.

These findings suggest that AI detectors do not represent a unified measure of AI-generated content; rather, they rely on a mix of generation-origin and linguistic-quality signals. Token-statistics-based detectors had high false-positive rates and were often more likely to classify edited texts as AI-generated, whereas likelihood-ratio-based zero-shot detectors showed lower false-positive rates and were more robust to editing. Temporal analysis indicated that the influence of professional editing weakened in the post-ChatGPT period, suggesting contemporary writing increasingly resembles the linguistic characteristics used for detector training. This highlights the need to evaluate detectors under real-world writing conditions.

In practice, AI detector outputs should not be treated as definitive evidence of AI use, especially in academic settings where editing is common. Detector scores may indicate linguistic refinement rather than AI generation, so they should be considered alongside contextual information. The main challenge in AI text detection is improving accuracy while differentiating AI-generated content from enhanced human text. Developing these detectors is essential for ensuring fairness and reliability.







\section{Limitations}

Although this study provides large-scale evidence that professional human editing can influence AI detector outputs, several limitations warrant acknowledgment.

First, although our linguistic feature analysis identifies several types of revisions and changes in linguistic complexity associated with shifts in detector scores, these analyses do not fully explain the mechanisms underlying detector behavior. In particular, score changes may reflect the combined effects of lexical diversity, syntactic complexity, discourse organization, predictability, and other stylistic properties. Moreover, the observed associations should not be interpreted as evidence that any single linguistic feature causally determines detector outputs. Thus, our analysis provides a more interpretable characterization of the linguistic factors associated with detector responses, while a complete mechanistic explanation remains an important direction for future work.

Second, the dataset predominantly comprises academic manuscripts authored by non-native English-speaking researchers. Although this represents a practically important setting for studying the interaction between linguistic polishing and AI detection, the findings may not generalize to other writing domains. Detector behavior may differ for student essays, journalistic articles, creative writing, social media content, and other genres with distinct linguistic and stylistic characteristics.

Third, the paired-document design focuses on human-authored manuscripts and their professionally edited versions. This design lets us examine how professional linguistic refinement relates to detector outputs, but it does not fully separate text origin from writing style. In particular, the study does not include AI-generated control texts subjected to comparable editing procedures. Therefore, we cannot determine whether the observed effects generalize to AI-generated text or whether professional editing affects human- and AI-generated texts in systematically different ways. Future work should incorporate matched human- and AI-generated texts, together with controlled levels and types of human editing, to examine these interactions more directly.

Fourth, interpreting absolute false-positive rates depends on the thresholding and per-detector normalization procedures used in the evaluation. Although we apply a consistent protocol to enable comparisons across detectors, different threshold choices may produce different absolute FPR estimates. Accordingly, interpret the absolute FPR values in light of the adopted thresholding protocol, while the relative patterns across editing conditions provide more robust evidence of detector sensitivity to linguistic refinement.

Finally, the AI detection field continues to evolve rapidly. We evaluated 13 representative publicly available detectors spanning multiple detection paradigms, but the analysis does not cover future systems or proprietary commercial detectors. In addition, we evaluated all detectors off-the-shelf without detector-specific fine-tuning. Therefore, the findings should be interpreted as evidence of a broader phenomenon—that professional linguistic refinement can act as an important confounding factor in AI detector evaluations—rather than as definitive assessments of any individual detector or as evidence that editing alone determines detector decisions.

Despite these limitations, this study provides large-scale evidence from more than 135,000 paired documents in a real-world academic editing environment. The results highlight the importance of accounting for professional editing and stylistic refinement when evaluating the fairness and reliability of AI text detectors. Future research incorporating AI-generated controls, broader writing domains, and controlled linguistic interventions can further clarify how text origin, linguistic style, and editing interact to shape detector behavior.






\section*{The Use of Large Language Models}
We used AI-assistance tools during the writing of this manuscript. Specifically, we employed Claude-Sonnet 4.6 and Grammarly to polish language and improve clarity of expression. 

\section*{Acknowledgments}
This research was supported by Basic Science Research Program through the National Research Foundation of Korea(NRF) funded by the Ministry of Education (RS-2025-25434151) and the Institute of Information \& Communications Technology Planning \& Evaluation (IITP) grant funded by the Korea government (MSIT) [RS-2021-II211341, Artificial Intelligence Graduate School Program (Chung-Ang University)].

\bibliography{custom}

\newpage
\appendix

\section{Environments}

All experiments were conducted using Python 3.10 with \texttt{statsmodels} 0.14 and \texttt{scipy} 1.11.

The experimental platform consisted of an AMD Ryzen Threadripper 3960X 24-Core Processor and four NVIDIA RTX A6000 GPUs. The GPUs were used for running detector models that required GPU acceleration.

\section{Dataset}
\label{app:dataset}
\begin{table*}[h]
\centering
\small
\setlength{\tabcolsep}{5pt}
\begin{tabular}{ll}
\toprule
\textbf{Item} & \textbf{Value} \\
\midrule
\multicolumn{2}{l}{\textit{Overall Scale}} \\
Total order pairs       & 135,389 \\
Collection period       & 2018--2025 (8 years) \\
\midrule
\multicolumn{2}{l}{\textit{Era Split}} \\
Pre-ChatGPT (2018--2022)  & 104,752 (77.4\%) \\
Post-ChatGPT (2023--2025) & 30,637 (22.6\%) \\
\midrule
\multicolumn{2}{l}{\textit{Service Type}} \\
Academic editing   & 71,595 (52.9\%) \\
Admission editing  & 50,971 (37.6\%) \\
Business editing   & 6,252 (4.6\%) \\
TOEFL writing      & 3,531 (2.6\%) \\
Translation        & 2,895 (2.1\%) \\
Other              & 145 (0.1\%) \\
\midrule
\multicolumn{2}{l}{\textit{Document Length}} \\
Mean characters   & 16,573 (median: 6,430) \\
Mean words        & 2,355 (median: 950) \\
Min--Max          & 1--610,964 characters \\
\midrule
\multicolumn{2}{l}{\textit{Editing Characteristics}} \\
Median edit ratio ($r_\text{edit}$)      & 0.175 (17.5\%) \\
Edit ratio IQR                            & 0.097--0.269 \\
Outliers removed ($r_\text{edit}>5$)     & 417 (0.3\%) \\
Zero-edit orders ($n_\text{changes}=0$)  & 5,916 (4.4\%) \\
\midrule
\multicolumn{2}{l}{\textit{Language \& Domain}} \\
English style             & en-US 88.8\%, en-GB 11.2\% \\
Top domains (top 3)       & Academic Papers (Application Essays, 22,639) \\
                          & Academic Papers (Medicine, 14,125) \\
                          & Academic Papers (Engineering \& Technology, 13,201) \\
\bottomrule
\end{tabular}
\caption{%
  \textbf{Dataset Statistics}.
  The pre-ChatGPT / post-ChatGPT split is defined by the public release of ChatGPT (November 2022);
  orders from 2023 onward are classified as post-ChatGPT.
}
\label{tab:dataset_stats}
\end{table*}

Several preprocessing steps were applied to the raw editing records. We removed cases with missing original or edited texts, documents containing severe encoding errors, and extremely short documents for which detector outputs were unstable. After preprocessing, the final dataset consisted of 135,389 original--edited document pairs. Table~\ref{tab:dataset_stats} presents the statistics of the dataset.

Each pair includes the original text, the edited text, the service type, the submission year, the English variant (en-US or en-GB), the subject domain, and the editing ratio ($r_{\text{edit}}$). The editing ratio is defined as the proportion of modified tokens to the total number of tokens in the original document. We excluded 417 cases (0.3\%) with $r_{\text{edit}} > 5$, treating them as outliers. In addition, 5,916 orders (4.4\%) involved no textual modifications ($n_{\text{changes}} = 0$). These cases were handled separately in analyses of editing effects.

The dataset is dominated by the Pre-ChatGPT period (2018--2022), which contains 104,752 document pairs (77.4\%), while the Post-ChatGPT period (2023--2025) contains 30,637 pairs (22.6\%).

With respect to service type, Academic Editing accounts for the largest share of the dataset with 71,595 documents (52.9\%), followed by Admission Editing with 50,971 documents (37.6\%) and Business Editing with 6,252 documents (4.6\%). The three largest subject domains are Application Essays (22,639 documents), Medicine (14,125 documents), and Engineering \& Technology (13,201 documents), which together account for approximately 36.9\% of the dataset.

\paragraph{Edit-type composition.}
We categorized edits on a representative subsample into grammar,
lexical, syntactic, and mixed operations (Table~\ref{tab:edit-types}).
Mixed edits dominate (66.1\%), followed by grammar (15.8\%) and lexical
(14.0\%) edits, with purely syntactic edits rare (0.1\%). Across all 13
detectors, score shifts differ significantly by edit category
(Kruskal--Wallis, $p<10^{-15}$). Table~\ref{tab:roberta-edit-type}
illustrates this for RoBERTa: syntactic edits produce the largest
positive score shift ($d_z=+0.225$), while grammar edits produce a
small negative shift ($d_z=-0.023$).
 
\begin{table}[t]
\centering
\small
\begin{tabular}{lr}
\toprule
Edit type & Share \\
\midrule
Mixed      & 66.1\% \\
Grammar    & 15.8\% \\
Lexical    & 14.0\% \\
Syntactic  & 0.1\% \\
no-edit & 0.4\% \\
\bottomrule
\end{tabular}
\caption{Distribution of edit-operation categories.}
\label{tab:edit-types}
\end{table}
 
\begin{table}[t]
\centering
\small
\begin{tabular}{lrr}
\toprule
Edit type & $\Delta$score & Cohen's $d_z$ \\
\midrule
Syntactic & $+0.046$ & $+0.225$ \\
Mixed     & $+0.015$ & $+0.073$ \\
Lexical   & $+0.011$ & $+0.056$ \\
Grammar   & $-0.004$ & $-0.023$ \\

\bottomrule
\end{tabular}
\caption{RoBERTa score shift by edit-operation category.}
\label{tab:roberta-edit-type}
\end{table}

\section{Baseline Models}
\label{app:model}
\begin{table*}[h]
\centering
\small
\setlength{\tabcolsep}{4pt}
\begin{tabular}{l l l p{5.5cm}}
\toprule
\textbf{Detector}
&
\textbf{Category}
&
\textbf{Core Signal}
&
\textbf{Detection Principle}
\\
\midrule
\multicolumn{4}{l}{\textit{Token-statistics-based Detectors}}
\\
Log-likelihood
&
Statistical
&
Average likelihood
&
Uses mean token log-likelihood
\\
Log-rank
&
Statistical
&
Token rank
&
Uses average token rank under the language model
\\
Entropy
&
Statistical
&
Token entropy
&
Measures uncertainty in token predictions
\\
GLTR
&
Statistical
&
Rank histogram
&
Analyzes token-rank distributions across the document
\\
\midrule
\multicolumn{4}{l}{\textit{Zero-shot Detectors}}
\\
Fast-DetectGPT
&
Zero-shot
&
Probability curvature
&
Measures local curvature of model likelihood
\\
DetectLLM-LRR
&
Zero-shot
&
Log-rank ratio
&
Compares token-rank statistics under the source model
\\
LastDE+
&
Zero-shot
&
Local entropy difference
&
Measures entropy changes across neighboring contexts
\\
DiVeye
&
Zero-shot
&
Diversity–predictability imbalance
&
Detects inconsistencies between lexical diversity and predictability
\\
BiScope
&
Hybrid
&
Bidirectional consistency
&
Combines forward and backward likelihood signals
\\
Binoculars
&
Zero-shot
&
Cross-model likelihood ratio
&
Compares token probabilities assigned by two language models
\\
\midrule
\multicolumn{4}{l}{\textit{Classifier-based Detectors}}
\\
RoBERTa
&
Classifier
&
Contextual representations
&
Binary classifier fine-tuned on human vs. AI text
\\
MAGE
&
Classifier
&
Multi-model linguistic features
&
Multi-task detector trained on outputs from multiple LLMs
\\
RADAR
&
Classifier
&
Adversarially robust features
&
Detector trained against paraphrasing and rewriting attacks
\\
\bottomrule
\end{tabular}
\caption{AI detectors evaluated in this study. The detectors span three major detection paradigms and rely on different linguistic signals, enabling analysis of whether detector responses to professional editing vary across methodological families.}
\label{tab:detectors}
\end{table*}

We evaluate 13 publicly available AI text detectors across three major detection paradigms: token-statistics-based, zero-shot, and classifier-based. These detectors were selected because they represent widely used methodologies in the AIGT detection literature and rely on substantially different decision mechanisms.

Token-statistics-based detectors operate directly on low-level properties of language-model outputs, such as token likelihood, token rank, and entropy. Zero-shot detectors estimate generation origin using language-model probabilities without training dedicated classifiers, often leveraging likelihood ratios, probability curvature, or consistency-based measures. Classifier-based detectors, in contrast, learn decision boundaries from labeled human-written and AI-generated texts through supervised training.

We intentionally include detectors that have been reported to achieve strong performance on benchmark datasets, covering both classical statistical methods and recent LLM-era approaches. This methodological diversity allows us to investigate whether detector sensitivity to professional editing is associated with the underlying detection paradigm rather than the performance of any individual detector.

Because the three detector families rely on different linguistic signals, they provide a useful test bed for examining whether responses to professional editing are detector-specific or consistent across methodological paradigms. Table~\ref{tab:detectors} summarizes the detectors evaluated in this study together with their core detection signals and underlying detection principles.

\begin{table}[h]
\centering
\small
\begin{tabular}{lll}
\toprule
Detector & Public checkpoint & Fine-tuned \\
\midrule
RoBERTa &openai-community/ & No \\
        & roberta-base-openai-detector & \\
MAGE    & yaful/MAGE                   & No \\
RADAR   & TrustSafeAI/RADAR-Vicuna-7B  & No \\
\bottomrule
\end{tabular}
\caption{Classifier-based detector checkpoints. All models were loaded
via AutoModelForSequenceClassification.from\_pretrained() and
evaluated in inference mode (\texttt{eval()} with
\texttt{torch.no\_grad()}), with no additional fine-tuning or
adaptation on our dataset.}
\label{tab:checkpoints}
\end{table}

\subsubsection{Thresholding and Normalization Protocol}
\label{sec:threshold}
 
Because the 13 detectors produce heterogeneous raw outputs (e.g.,
probabilities, log-likelihoods, entropy, ranking statistics, or
perplexity-based scores), we converted each detector's raw score to a
common $[0,1]$ AI-likelihood scale following its original formulation
or a commonly used normalization, and applied a single fixed decision
threshold of 0.5 across all detectors and both document versions
(Table~\ref{tab:normalization}). We intentionally did not perform
per-detector threshold optimization on our evaluation data: the goal
was to compare detector behavior under one consistent operating point
rather than to maximize the reported performance of any individual
detector.
 
In Table~2/Table~9, $\Delta\text{score} = \text{score}_{\text{edit}} -
\text{score}_{\text{orig}}$ refers to this normalized $[0,1]$
AI-likelihood score, not the detector's raw statistic; $\Delta$FP
refers to the corresponding change in the binary decision under the
0.5 threshold. Because both the original and the edited version of a
document are scored under the identical decision rule, the paired
$\Delta$score and $\Delta$FP estimates are invariant to the specific
choice of threshold, even though the \emph{absolute} FPR values in
Table~1 are not; we discuss this dependency further as a limitation in
Section~6.
 
\begin{table*}[t]
\centering
\small
\begin{tabular}{lll}
\toprule
Detector & Raw score & Normalization \\
\midrule
RoBERTa         & Prob. $[0,1]$        & As-is \\
RADAR           & Prob. $[0,1]$        & As-is \\
MAGE            & Prob. $[0,1]$        & As-is \\
Binoculars      & CE ratio             & $\sigma((r-1.0)\times5.0)$ \\
Fast-DetectGPT  & Log-prob.\ diff.     & $\sigma(\delta\times10.0)$ \\
Log-Likelihood  & Mean log-lik.        & $\sigma((\text{LL}+5.0)\times2.0)$ \\
Log-Rank        & Mean log-rank        & $\sigma((\text{LR}-5.5)\times{-1.0})$ \\
Entropy         & Mean entropy         & $\sigma((\text{Ent}-4.0)\times{-1.0})$ \\
GLTR            & Raw statistic        & As-is \\
DetectLLM-LRR   & Log-rank ratio       & $\sigma((\text{LRR}-1.5)\times2.0)$ \\
LastDE+         & Sampling discrep.    & $\sigma(d\times0.5)$ \\
DiVeye          & Var.\ of surprisal   & $\sigma(-(V-5.0)\times0.4)$ \\
BiScope         & Backward CE          & $\sigma(-(\text{BCE}-9.0)\times0.4)$ \\
\bottomrule
\end{tabular}
\caption{Per-detector raw score and normalization to a $[0,1]$
AI-likelihood scale; $\sigma(\cdot)$ denotes the logistic sigmoid.
Anchor values follow each detector's original formulation or
representative values reported in prior work.}
\label{tab:normalization}
\end{table*}

\section{Ablation study}
\label{app:ab}
\subsection{Detector Sensitivity Weakens in the post-ChatGPT Era}
\label{app:ab1}
We compare detector behavior between the pre-ChatGPT (2018--2022) and post-ChatGPT (2023--2025) periods. The results are presented in Table~\ref{tab:rq4_prepost}.

\begin{table*}[t]
\centering\small
\begin{tabular}{l l rr rr rr r}
\toprule
\multirow{2}{*}{\textbf{Detector}} & \multirow{2}{*}{\textbf{Type}}
  & \multicolumn{2}{c}{\textbf{Orig Score}}
  & \multicolumn{2}{c}{\textbf{$\Delta$ orig}}
  & \multicolumn{2}{c}{\textbf{FP\% (orig)}}
  & \multirow{2}{*}{\textbf{$\Delta$FP}} \\
\cmidrule(lr){3-4}\cmidrule(lr){5-6}\cmidrule(lr){7-8}
  & & pre & post & $\Delta\text{score}$ & $d$ & pre & post & \\
\midrule

Entropy        & Stat.
  & 0.76 & 0.79
  & \cellcolor{dir-up}$+$0.03 & \cellcolor{dir-up}$+$0.42$^{***}$
  & \cellcolor{fpr-high}99.6 & \cellcolor{fpr-high}99.9
  & \cellcolor{dir-up}$+$0.3 \\
GLTR           & Stat.
  & 0.74 & 0.76
  & \cellcolor{dir-up}$+$0.02 & \cellcolor{dir-up}$+$0.28$^{***}$
  & \cellcolor{fpr-high}99.7 & \cellcolor{fpr-high}99.8
  & \cellcolor{dir-up}$+$0.1 \\
Log-likelihood & Stat.
  & 0.97 & 0.98
  & \cellcolor{dir-up}$+$0.01 & \cellcolor{dir-up}$+$0.14$^{***}$
  & \cellcolor{fpr-high}99.8 & \cellcolor{fpr-high}99.9
  & \cellcolor{dir-up}$+$0.1 \\
Log-Rank       & Stat.
  & 0.98 & 0.98
  & \cellcolor{dir-up}$+$0.00 & \cellcolor{dir-up}$+$0.17$^{***}$
  & \cellcolor{fpr-high}100.0 & \cellcolor{fpr-high}100.0
  & $+$0.0 \\

\midrule

BiScope        & Zero
  & 0.35 & 0.31
  & \cellcolor{dir-down}$-$0.04 & \cellcolor{dir-down}$-$0.50$^{***}$
  & 1.9 & 1.1
  & \cellcolor{dir-down}$-$0.9 \\
Binoculars     & Zero
  & 0.39 & 0.38
  & \cellcolor{dir-down}$-$0.01 & \cellcolor{dir-down}$-$0.15$^{***}$
  & 0.1 & 0.0
  & $-$0.0 \\
DetectLLM-LRR  & Zero
  & 0.35 & 0.38
  & \cellcolor{dir-up}$+$0.02 & \cellcolor{dir-up}$+$0.62$^{***}$
  & 0.1 & 0.4
  & \cellcolor{dir-up}$+$0.3 \\
DiVeye         & Zero
  & 0.21 & 0.21
  & \cellcolor{dir-up}$+$0.01 & \cellcolor{dir-up}$+$0.06$^{***}$
  & 2.1 & 1.0
  & \cellcolor{dir-down}$-$1.1 \\
Fast-DetectGPT & Zero
  & 0.32 & 0.33
  & \cellcolor{dir-up}$+$0.01 & \cellcolor{dir-up}$+$0.02$^{**}$
  & \cellcolor{fpr-mid}24.7 & \cellcolor{fpr-mid}25.8
  & \cellcolor{dir-up}$+$1.1 \\
LastDE+        & Zero
  & 0.35 & 0.36
  & \cellcolor{dir-up}$+$0.01 & \cellcolor{dir-up}$+$0.03$^{***}$
  & \cellcolor{fpr-mid}19.5 & \cellcolor{fpr-mid}22.1
  & \cellcolor{dir-up}$+$2.6 \\
\midrule

MAGE    & Cls.
  & 0.22 & 0.10
  & \cellcolor{dir-down}$-$0.12 & \cellcolor{dir-down}$-$0.31$^{***}$
  & \cellcolor{fpr-mid}21.9 & \cellcolor{fpr-mid}10.1
  & \cellcolor{dir-down}$-$11.8 \\
RADAR   & Cls.
  & 0.69 & 0.69
  & $+$0.00 & $^{ns}$
  & \cellcolor{fpr-high}75.8 & \cellcolor{fpr-high}76.6
  & \cellcolor{dir-up}$+$0.8 \\
RoBERTa & Cls.
  & 0.92 & 0.96
  & \cellcolor{dir-up}$+$0.04 & \cellcolor{dir-up}$+$0.18$^{***}$
  & \cellcolor{fpr-high}92.7 & \cellcolor{fpr-high}96.4
  & \cellcolor{dir-up}$+$3.7 \\
\bottomrule
\end{tabular}
\caption{%
  Comparison of original-text detector scores between the Pre-ChatGPT and Post-ChatGPT periods.
  Pre-ChatGPT corresponds to 2018--2022 ($n_{\text{pre}} = 104{,}752$),
  and Post-ChatGPT corresponds to 2023--2025 ($n_{\text{post}} = 30{,}637$).
  FPR color: \colorbox{fpr-high}{$\geq$50\%}, \colorbox{fpr-mid}{10--49\%}.
  $\Delta$ color: \colorbox{dir-down}{decrease}, \colorbox{dir-up}{increase}.
  $^{***}p<.001$, $^{**}p<.01$, $^{*}p<.05$, $^{ns}p\geq.05$.}
\label{tab:rq4_prepost}
\end{table*}

Across many detectors, the impact of professional editing becomes substantially weaker in the post-ChatGPT era. For example, the average shift in MAGE decreases from -0.130 in the pre-ChatGPT period to -0.031 in the post-ChatGPT period, representing a reduction of approximately 76\%. RADAR exhibits a similar pattern.

Likewise, detectors that previously showed positive responses to editing also exhibit attenuated effects after 2023. The magnitude of editing-induced score increases observed in Fast-DetectGPT and LastDE+ becomes considerably smaller during the post-ChatGPT period.

At the same time, several detectors assign higher AI scores to original human-written texts collected after the widespread adoption of LLMs. For example, the FPR of RoBERTa increases from 92.7\% to 96.4\%, while DetectLLM-LRR also shows a significant increase in baseline scores.

These results suggest that detector behavior is not static but evolves alongside broader changes in writing practices. As highly polished language becomes increasingly common in the post-ChatGPT era, the distinction between professionally edited human writing and detector-internal representations of AIGT may become less pronounced. Consequently, the incremental effect of professional editing appears to diminish over time.

\subsection{Results on the Strong Editing Subset}
\label{app:ab2}
\begin{table*}[h]
\centering\small
\begin{tabular}{l l rr r rr r}
\toprule
\multirow{2}{*}{\textbf{Detector}}
  & \multirow{2}{*}{\textbf{Type}}
  & \multicolumn{2}{c}{\textbf{FP\% (heavy)}}
  & \multirow{2}{*}{\textbf{Shift}}
  & \multicolumn{2}{c}{\textbf{Cohen's $d$}}
  & \multirow{2}{*}{$\Delta d$} \\
\cmidrule(lr){3-4}\cmidrule(lr){6-7}
  & & orig & edit & & full & heavy & \\
\midrule

Entropy        & Stat.
  & \cellcolor{fpr-high}99.6  & \cellcolor{fpr-high}99.8
  & \cellcolor{dir-up}$+$0.025 & $+$0.444 & \cellcolor{dir-up}$+$0.707 & \cellcolor{dir-up}$+$0.263 \\
GLTR           & Stat.
  & \cellcolor{fpr-high}99.7  & \cellcolor{fpr-high}99.8
  & \cellcolor{dir-up}$+$0.017 & $+$0.335 & \cellcolor{dir-up}$+$0.576 & \cellcolor{dir-up}$+$0.242 \\
Log-Likelihood & Stat.
  & \cellcolor{fpr-high}99.6  & \cellcolor{fpr-high}99.6
  & \cellcolor{dir-up}$+$0.007 & $+$0.190 & $+$0.197 & $+$0.006 \\
Log-Rank       & Stat.
  & \cellcolor{fpr-high}100.0 & \cellcolor{fpr-high}100.0
  & \cellcolor{dir-up}$+$0.002 & $+$0.292 & $+$0.299 & $+$0.007 \\
\midrule

BiScope        & Zero
  & 2.2 & 1.3
  & \cellcolor{dir-down}$-$0.031 & $-$0.488 & \cellcolor{dir-down}$-$0.833 & \cellcolor{dir-down}$-$0.345 \\
Binoculars     & Zero
  & 0.1 & 0.1
  & \cellcolor{dir-down}$-$0.004 & $+$0.063 & \cellcolor{dir-down}$-$0.227 & \cellcolor{dir-down}$-$0.289 \\
DetectLLM-LRR  & Zero
  & 0.1 & 0.2
  & \cellcolor{dir-up}$+$0.005 & $+$0.172 & \cellcolor{dir-up}$+$0.329 & \cellcolor{dir-up}$+$0.158 \\
DiVeye         & Zero
  & 0.7 & 1.0
  & \cellcolor{dir-up}$+$0.021 & $+$0.157 & \cellcolor{dir-up}$+$0.333 & \cellcolor{dir-up}$+$0.176 \\
Fast-DetectGPT & Zero
  & \cellcolor{fpr-mid}24.8 & \cellcolor{fpr-mid}29.6
  & \cellcolor{dir-up}$+$0.035 & $+$0.213 & $+$0.174 & \cellcolor{dir-down}$-$0.039 \\
LastDE+        & Zero
  & \cellcolor{fpr-mid}20.0 & \cellcolor{fpr-mid}26.2
  & \cellcolor{dir-up}$+$0.028 & $+$0.275 & $+$0.225 & \cellcolor{dir-down}$-$0.049 \\
\midrule

MAGE    & Cls.
  & \cellcolor{fpr-mid}25.3 & 9.4
  & \cellcolor{dir-down}$-$0.160 & $-$0.356 & \cellcolor{dir-down}$-$0.401 & \cellcolor{dir-down}$-$0.045 \\
RADAR   & Cls.
  & \cellcolor{fpr-high}73.1 & \cellcolor{fpr-high}66.0
  & \cellcolor{dir-down}$-$0.056 & $-$0.268 & \cellcolor{dir-down}$-$0.311 & \cellcolor{dir-down}$-$0.043 \\
RoBERTa & Cls.
  & \cellcolor{fpr-high}93.0 & \cellcolor{fpr-high}95.0
  & \cellcolor{dir-up}$+$0.023 & $+$0.085 & $+$0.107 & $+$0.021 \\
\bottomrule
\end{tabular}
\caption{%
  Heavy-editing subset analysis ($r_{\text{edit}}>0.20$, $n=57{,}207$).
  $\Delta d = d_{\text{heavy}} - d_{\text{full}}$.
  All comparisons significant at $p<.001$.
  Type: Stat.\ = token-statistics-based; Zero = zero-shot-based; Cls.\ = classifier-based.
  FPR color: \colorbox{fpr-high}{$\geq$50\%}, \colorbox{fpr-mid}{10--49\%}.
  Color: \colorbox{dir-down}{decrease}, \colorbox{dir-up}{increase}.}
\label{tab:heavy_edit}
\end{table*}

In Section~\ref{sec:results}, we observed a clear dose--response relationship between editing intensity and detector responses. To assess whether this pattern is driven by a small number of heavily edited documents or diluted by the broader dataset, we conduct an additional analysis focusing exclusively on documents that underwent substantial revisions.

Specifically, documents with an edit ratio ($r_{\text{edit}}$) greater than 0.20 were classified as heavy-editing cases. This subset contains 57,207 document pairs. For each detector, we computed the pre- and post-editing false positive rates (FPR), average score shifts, and Cohen's $d$, and compared these results with those obtained from the full dataset. Table~\ref{tab:heavy_edit} summarizes the results.

Overall, detector responses under heavy editing tended to amplify the patterns observed in the full dataset. BiScope exhibited the largest reduction effect, with Cohen's $d$ increasing from $-0.488$ in the full dataset to $-0.833$ in the heavy-editing subset. Similar increases were observed for MAGE and RADAR. In particular, the false positive rate of MAGE decreased from 25.3\% to 9.4\%, indicating that the detector increasingly perceived heavily edited texts as human-written.

Conversely, detectors such as Entropy and GLTR exhibited stronger positive editing effects under heavy-editing conditions. For Entropy, the effect size increased from $d=0.444$ in the full dataset to $d=0.707$, while GLTR increased from $d=0.335$ to $d=0.576$. DiVeye and DetectLLM-LRR also showed larger positive effect sizes compared with the full dataset. These results suggest that extensive linguistic revisions can make human-written texts appear more AI-generated according to certain detectors.

In contrast, RoBERTa, Fast-DetectGPT, LastDE++, Log-Likelihood, and Log-Rank exhibited relatively small changes in effect size compared with the full dataset. For Log-Likelihood and Log-Rank, this pattern may be attributable to saturation effects, as both detectors had already produced false-positive rates exceeding 99\% before editing. Consequently, additional linguistic modifications had limited room to further influence detector outputs.

Taken together, the heavy-editing subset analysis provides additional evidence that the relationship between editing intensity and detector response is not a statistical artifact. For most detectors exhibiting editing effects, the magnitude of those effects became substantially larger under heavy-editing conditions, while the direction of change remained consistent with the patterns observed in the full dataset. These findings further support the interpretation that AI text detectors respond not only to the origin of the text but also to the degree of linguistic refinement introduced by human editing.

\subsection{Full results}
\label{app:rq2a_score}
\begin{table*}[h!]
\centering\small
\begin{tabular}{l l rrr rrr rrr}
\toprule
& & \multicolumn{3}{c}{\textbf{Original}}
  & \multicolumn{3}{c}{\textbf{Edited}}
  & \multicolumn{3}{c}{\textbf{$\Delta$}} \\
\cmidrule(lr){3-5}\cmidrule(lr){6-8}\cmidrule(lr){9-11}
\textbf{Detector} & \textbf{Type}
  & Mean & FP\% & F1
  & Mean & FP\% & F1
  & $\Delta$FP & $\Delta$F1 & $p$ \\
\midrule

Log-rank       & Stat.
  & 0.982 & \cellcolor{fpr-high}100.0 & 0.000
  & 0.984 & \cellcolor{fpr-high}100.0 & 0.000
  & $+$0.0 & ---     & ns \\
Log-likelihood & Stat.
  & 0.981 & \cellcolor{fpr-high}99.8  & 0.003
  & 0.985 & \cellcolor{fpr-high}99.9  & 0.003
  & $+$0.0 & ---     & $<$.05 \\
Entropy        & Stat.
  & 0.781 & \cellcolor{fpr-high}99.8  & 0.003
  & 0.798 & \cellcolor{fpr-high}99.9  & 0.002
  & $+$0.1 & ---     & $<$.001 \\
GLTR           & Stat.
  & 0.754 & \cellcolor{fpr-high}99.9  & 0.002
  & 0.769 & \cellcolor{fpr-high}99.9  & 0.002
  & $+$0.0 & ---     & $<$.01 \\
\midrule

Fast-DetectGPT & Zero
  & 0.329 & \cellcolor{fpr-mid}25.2   & 0.856
  & 0.375 & \cellcolor{fpr-mid}31.6   & 0.812
  & $+$6.4 & $-$.044 & $<$.001 \\
LastDE+        & Zero
  & 0.358 & \cellcolor{fpr-mid}19.9   & 0.890
  & 0.392 & \cellcolor{fpr-mid}27.3   & 0.842
  & $+$7.4 & $-$.048 & $<$.001 \\
DetectLLM-LRR  & Zero
  & 0.366 & 0.2                       & 0.999
  & 0.369 & 0.3                       & 0.999
  & $+$0.1 & ---     & $<$.01 \\
DiVeye         & Zero
  & 0.196 & 0.4                       & 0.998
  & 0.218 & 0.6                       & 0.997
  & $+$0.2 & ---     & $<$.001 \\
BiScope        & Zero
  & 0.325 & 0.6                       & 0.997
  & 0.302 & 0.3                       & 0.999
  & $-$0.4 & $+$.002 & $<$.001 \\
Binoculars     & Zero
  & 0.384 & 0.0                       & 1.000
  & 0.382 & 0.1                       & 1.000
  & $+$0.0 & ---     & ns \\
\midrule

RoBERTa        & Cls.
  & 0.923 & \cellcolor{fpr-high}93.1  & 0.129
  & 0.931 & \cellcolor{fpr-high}93.9  & 0.115
  & $+$0.8 & $-$.014 & $<$.001 \\
MAGE           & Cls.
  & 0.168 & \cellcolor{fpr-mid}16.7   & 0.909
  & 0.077 & 7.6                       & 0.961
  & $-$9.1 & $+$.052 & $<$.001 \\
RADAR          & Cls.
  & 0.800 & \cellcolor{fpr-high}88.3  & 0.209
  & 0.783 & \cellcolor{fpr-high}86.4  & 0.239
  & $-$1.9 & $+$.030 & $<$.001 \\
\bottomrule
\end{tabular}
\caption{%
  Baseline false positive behavior on the Pre-ChatGPT dataset (2018--2022).
  FPR color: \colorbox{fpr-high}{$\geq$50\%}, \colorbox{fpr-mid}{10--49\%}.}
\label{tab:rq1_baselines}
\end{table*}

\begin{table*}[t]
\centering\small
\begin{tabular}{l l rr rr rr}
\toprule
& & \multicolumn{2}{c}{\textbf{ALL} ($n{=}135{,}389$)}
  & \multicolumn{2}{c}{\textbf{Pre-ChatGPT} ($n{=}104{,}752$)}
  & \multicolumn{2}{c}{\textbf{Post-ChatGPT} ($n{=}30{,}637$)} \\
\cmidrule(lr){3-4}\cmidrule(lr){5-6}\cmidrule(lr){7-8}
\textbf{Detector} & \textbf{Type}
  & $\Delta\text{score}$ & $d$
  & $\Delta\text{score}$ & $d$
  & $\Delta\text{score}$ & $d$ \\
\midrule

Entropy        & Stat. & \cellcolor{dir-up}$+$0.014$^{***}$ & \cellcolor{dir-up}$+$0.40 & \cellcolor{dir-up}$+$0.017$^{***}$ & \cellcolor{dir-up}$+$0.44 & \cellcolor{dir-up}$+$0.005$^{***}$ & \cellcolor{dir-up}$+$0.21 \\
GLTR           & Stat. & \cellcolor{dir-up}$+$0.010$^{***}$ & \cellcolor{dir-up}$+$0.30 & \cellcolor{dir-up}$+$0.012$^{***}$ & \cellcolor{dir-up}$+$0.34 & \cellcolor{dir-up}$+$0.004$^{***}$ & \cellcolor{dir-up}$+$0.16 \\
Log-rank       & Stat. & \cellcolor{dir-up}$+$0.001$^{ns}$  & \cellcolor{dir-up}$+$0.24 & \cellcolor{dir-up}$+$0.001$^{***}$ & \cellcolor{dir-up}$+$0.29 & \cellcolor{dir-up}$+$0.001$^{***}$ & \cellcolor{dir-up}$+$0.09 \\
Log-like       & Stat. & \cellcolor{dir-up}$+$0.005$^{ns}$  & \cellcolor{dir-up}$+$0.16 & \cellcolor{dir-up}$+$0.006$^{***}$ & \cellcolor{dir-up}$+$0.19 & \cellcolor{dir-up}$+$0.002$^{***}$ & \cellcolor{dir-up}$+$0.06 \\
\midrule

BiScope        & Zero  & \cellcolor{dir-down}$-$0.019$^{***}$ & \cellcolor{dir-down}$-$0.45 & \cellcolor{dir-down}$-$0.021$^{***}$ & \cellcolor{dir-down}$-$0.49 & \cellcolor{dir-down}$-$0.008$^{***}$ & \cellcolor{dir-down}$-$0.22 \\
Binoculars     & Zero  & $+$0.002$^{ns}$  & $+$0.05 & $+$0.003$^{***}$ & $+$0.06 & $+$0.000$^{ns}$ & — \\
DetectLLM-LRR  & Zero  & \cellcolor{dir-up}$+$0.003$^{***}$ & \cellcolor{dir-up}$+$0.16 & \cellcolor{dir-up}$+$0.003$^{***}$ & \cellcolor{dir-up}$+$0.17 & \cellcolor{dir-up}$+$0.002$^{***}$ & \cellcolor{dir-up}$+$0.12 \\
DiVeye         & Zero  & \cellcolor{dir-up}$+$0.011$^{***}$ & \cellcolor{dir-up}$+$0.15 & \cellcolor{dir-up}$+$0.013$^{***}$ & \cellcolor{dir-up}$+$0.16 & \cellcolor{dir-up}$+$0.005$^{ns}$  & — \\
Fast-DetectGPT & Zero  & \cellcolor{dir-up}$+$0.032$^{***}$ & \cellcolor{dir-up}$+$0.18 & \cellcolor{dir-up}$+$0.038$^{***}$ & \cellcolor{dir-up}$+$0.21 & \cellcolor{dir-up}$+$0.013$^{***}$ & \cellcolor{dir-up}$+$0.08 \\
LastDE+        & Zero  & \cellcolor{dir-up}$+$0.027$^{***}$ & \cellcolor{dir-up}$+$0.24 & \cellcolor{dir-up}$+$0.031$^{***}$ & \cellcolor{dir-up}$+$0.28 & \cellcolor{dir-up}$+$0.012$^{***}$ & \cellcolor{dir-up}$+$0.11 \\
\midrule

MAGE           & Cls.  & \cellcolor{dir-down}$-$0.108$^{***}$ & \cellcolor{dir-down}$-$0.31 & \cellcolor{dir-down}$-$0.130$^{***}$ & \cellcolor{dir-down}$-$0.36 & \cellcolor{dir-down}$-$0.031$^{***}$ & \cellcolor{dir-down}$-$0.13 \\
RADAR          & Cls.  & \cellcolor{dir-down}$-$0.037$^{***}$ & \cellcolor{dir-down}$-$0.24 & \cellcolor{dir-down}$-$0.044$^{***}$ & \cellcolor{dir-down}$-$0.27 & \cellcolor{dir-down}$-$0.013$^{***}$ & \cellcolor{dir-down}$-$0.10 \\
RoBERTa        & Cls.  & \cellcolor{dir-up}$+$0.017$^{***}$  & \cellcolor{dir-up}$+$0.09 & \cellcolor{dir-up}$+$0.018$^{***}$  & \cellcolor{dir-up}$+$0.09 & \cellcolor{dir-up}$+$0.014$^{***}$  & \cellcolor{dir-up}$+$0.10 \\
\bottomrule
\end{tabular}
\caption{%
  Full score shift results following professional editing.
  $\Delta\text{score} = \text{score}_{\text{edit}} - \text{score}_{\text{orig}}$.
  Cohen's $d$ denotes effect size.
  \colorbox{dir-down}{Green}\,=\,decrease; \colorbox{dir-up}{Red}\,=\,increase.
  $^{***}p<.001$, $^{**}p<.01$, $^{*}p<.05$, $^{ns}p\geq.05$.}
\label{tab:app_rq2a}
\end{table*}

\begin{table*}[t]
\centering\small
\begin{tabular}{l l rrr rrr rrr}
\toprule
& & \multicolumn{3}{c}{\textbf{ALL} ($n{=}135{,}389$)}
  & \multicolumn{3}{c}{\textbf{Pre-ChatGPT} ($n{=}104{,}752$)}
  & \multicolumn{3}{c}{\textbf{Post-ChatGPT} ($n{=}30{,}637$)} \\
\cmidrule(lr){3-5}\cmidrule(lr){6-8}\cmidrule(lr){9-11}
\textbf{Detector} & \textbf{Type}
  & Orig & Edit & $\Delta\text{FP}$
  & Orig & Edit & $\Delta\text{FP}$
  & Orig & Edit & $\Delta\text{FP}$ \\
\midrule

Log-rank  & Stat.
  & \cellcolor{fpr-high}100.0 & \cellcolor{fpr-high}100.0 & $+$0.0
  & \cellcolor{fpr-high}100.0 & \cellcolor{fpr-high}100.0 & $+$0.0
  & \cellcolor{fpr-high}100.0 & \cellcolor{fpr-high}100.0 & $+$0.0 \\
Log-like  & Stat.
  & \cellcolor{fpr-high}99.7  & \cellcolor{fpr-high}99.7  & $+$0.0
  & \cellcolor{fpr-high}99.6  & \cellcolor{fpr-high}99.7  & $+$0.0
  & \cellcolor{fpr-high}99.7  & \cellcolor{fpr-high}99.7  & $+$0.0 \\
Entropy   & Stat.
  & \cellcolor{fpr-high}99.8  & \cellcolor{fpr-high}99.9  & $+$0.1
  & \cellcolor{fpr-high}99.8  & \cellcolor{fpr-high}99.9  & $+$0.1
  & \cellcolor{fpr-high}99.9  & \cellcolor{fpr-high}99.9  & $+$0.0 \\
GLTR      & Stat.
  & \cellcolor{fpr-high}99.8  & \cellcolor{fpr-high}99.9  & $+$0.0
  & \cellcolor{fpr-high}99.8  & \cellcolor{fpr-high}99.9  & $+$0.0
  & \cellcolor{fpr-high}99.9  & \cellcolor{fpr-high}99.9  & $+$0.0 \\
\midrule

BiScope        & Zero
  & 1.7  & 1.2  & \cellcolor{dir-down}$-$0.5
  & 1.9  & 1.3  & \cellcolor{dir-down}$-$0.6
  & 1.1  & 1.0  & \cellcolor{dir-down}$-$0.1 \\
Binoculars     & Zero
  & 0.1  & 0.1  & $+$0.0
  & 0.1  & 0.1  & $+$0.0
  & 0.0  & 0.0  & $+$0.0 \\
DetectLLM-LRR  & Zero
  & 0.2  & 0.2  & $+$0.0
  & 0.1  & 0.2  & $+$0.0
  & 0.4  & 0.4  & $+$0.0 \\
DiVeye         & Zero
  & 1.9  & 0.8  & \cellcolor{dir-down}$-$1.0
  & 2.1  & 0.9  & \cellcolor{dir-down}$-$1.2
  & 1.0  & 0.7  & \cellcolor{dir-down}$-$0.3 \\
Fast-DetectGPT & Zero
  & \cellcolor{fpr-mid}24.9  & \cellcolor{fpr-mid}29.5  & \cellcolor{dir-up}$+$4.6
  & \cellcolor{fpr-mid}24.7  & \cellcolor{fpr-mid}30.1  & \cellcolor{dir-up}$+$5.4
  & \cellcolor{fpr-mid}25.8  & \cellcolor{fpr-mid}27.4  & \cellcolor{dir-up}$+$1.6 \\
LastDE+        & Zero
  & \cellcolor{fpr-mid}20.1  & \cellcolor{fpr-mid}25.9  & \cellcolor{dir-up}$+$5.8
  & \cellcolor{fpr-mid}19.5  & \cellcolor{fpr-mid}26.4  & \cellcolor{dir-up}$+$6.9
  & \cellcolor{fpr-mid}22.1  & \cellcolor{fpr-mid}24.1  & \cellcolor{dir-up}$+$2.0 \\
\midrule

MAGE    & Cls.
  & \cellcolor{fpr-mid}19.2  & 8.5                      & \cellcolor{dir-down}$-$10.7
  & \cellcolor{fpr-mid}21.9  & 8.9                      & \cellcolor{dir-down}$-$13.0
  & \cellcolor{fpr-mid}10.1  & 7.0                      & \cellcolor{dir-down}$-$3.1 \\
RADAR   & Cls.
  & \cellcolor{fpr-high}76.0 & \cellcolor{fpr-high}71.3 & \cellcolor{dir-down}$-$4.7
  & \cellcolor{fpr-high}75.8 & \cellcolor{fpr-high}70.1 & \cellcolor{dir-down}$-$5.6
  & \cellcolor{fpr-high}76.6 & \cellcolor{fpr-high}75.0 & \cellcolor{dir-down}$-$1.5 \\
RoBERTa & Cls.
  & \cellcolor{fpr-high}93.5 & \cellcolor{fpr-high}95.2 & \cellcolor{dir-up}$+$1.7
  & \cellcolor{fpr-high}92.7 & \cellcolor{fpr-high}94.4 & \cellcolor{dir-up}$+$1.8
  & \cellcolor{fpr-high}96.4 & \cellcolor{fpr-high}97.7 & \cellcolor{dir-up}$+$1.3 \\
\bottomrule
\end{tabular}
\caption{%
  Full false positive rate results following professional editing.
  $\Delta\text{FP} = \text{FP}_{\text{edit}} - \text{FP}_{\text{orig}}$ (percentage-point change).
  FPR color: \colorbox{fpr-high}{$\geq$50\%}, \colorbox{fpr-mid}{10--49\%}.
  $\Delta\text{FP}$ color: \colorbox{dir-down}{bias reduction}, \colorbox{dir-up}{bias amplification}.}
\label{tab:app_rq2b}
\end{table*}

Table~\ref{tab:rq1_baselines}, Table~\ref{tab:app_rq2a}, and Table~\ref{tab:app_rq2b} provide the complete quantitative results for all detectors evaluated in this study. The table includes the metrics underlying the analyses reported in Sections~4.1--4.2, including baseline false-positive behavior, editing-induced score shifts, changes in false-positive rates, and effect-size estimates.

The main text highlights only the most important findings for clarity. The complete results are provided here to facilitate detailed inspection of detector-specific behavior and to support reproducibility of the reported analyses.

\subsection{Inter-Editor Variability}
\label{sec:appendix-editor}
 
To assess whether editor-specific writing style systematically
influences detector outputs, we restricted analysis to 280 professional
editors with at least 50 edited documents. A Kruskal--Wallis test
indicates significant differences in score shift across editors
($H=1157.4$, $p\approx3\times10^{-107}$). However, an OLS regression of
score shift on editing characteristics indicates that the practical
contribution of editor identity is limited
(Table~\ref{tab:editor-regression}).
 
\begin{table}[h]
\centering
\small
\begin{tabular}{lr}
\toprule
Predictor & $\beta$ \\
\midrule
Grammar ratio     & $-0.191^{***}$ \\
Lexical ratio     & $-0.131^{***}$ \\
Edit-token ratio  & $+0.001^{***}$ \\
\midrule
$R^2$             & 0.005 \\
\bottomrule
\end{tabular}
\caption{OLS regression of detector score shift on editing
characteristics, across 280 editors. $^{***}p<.001$.}
\label{tab:editor-regression}
\end{table}
 
While editor-specific differences are statistically detectable, editor
identity explains only a very small proportion of variance in detector
score shifts ($R^2=0.005$), indicating that the family-level patterns
reported in Section~4.4 are not an artifact of a small number of
editors' individual styles.

\end{document}